\pdfoutput=1

\documentclass[11pt]{article}

\usepackage[]{ACL2023}

\usepackage{times}
\usepackage{latexsym}

\usepackage[T1]{fontenc}

\usepackage[utf8]{inputenc}

\usepackage{microtype}

\usepackage{inconsolata}

\usepackage{tabularx}
\usepackage{graphicx}
\usepackage{adjustbox}
\usepackage{multirow}
\usepackage{booktabs}
\usepackage{bbding}
\usepackage{pifont}
\usepackage{wasysym}
\usepackage{amssymb}
\usepackage{arydshln}
\usepackage{caption}
\usepackage{subcaption}
\usepackage{relsize}
\usepackage{tabto}
\usepackage{siunitx}
\usepackage{arydshln}
\usepackage{cleveref}
\usepackage{xcolor}
\usepackage{courier}
\usepackage[normalem]{ulem}

\crefformat{section}{\S#2#1#3}
\crefformat{subsection}{\S#2#1#3}
\crefformat{subsubsection}{\S#2#1#3}

\title{\textsc{DEnsity}: Open-domain Dialogue Evaluation Metric \\ using Density Estimation}

\author{ChaeHun Park \hspace{0.3cm} Seungil Chad Lee \hspace{0.3cm} Daniel Rim \hspace{0.3cm} \textbf{Jaegul Choo}\\
KAIST AI \\
\hspace{0cm}\texttt{\{ddehun,silly5921,ssong88,jchoo\}@kaist.ac.kr} \\
 }

\begin{document}
\maketitle
\begin{abstract}

Despite the recent advances in open-domain dialogue systems, building a reliable evaluation metric is still a challenging problem. 
Recent studies proposed learnable metrics based on classification models trained to distinguish the correct response.
However, neural classifiers are known to make overly confident predictions for examples from unseen distributions.
We propose \textsc{DEnsity}, which evaluates a response by utilizing density estimation on the feature space derived from a neural classifier. 
Our metric measures how likely a response would appear in the distribution of human conversations.
Moreover, to improve the performance of \textsc{DEnsity}, we utilize contrastive learning to further compress the feature space.
Experiments on multiple response evaluation datasets show that \textsc{DEnsity} correlates better with human evaluations than the existing metrics.\footnote{Our code is available at \url{https://github.com/ddehun/DEnsity}.}

\end{abstract}

\section{Introduction}
Automatic evaluation is essential in developing various natural language generation systems, such as machine translation~\citep{sutskever2014sequence} and summarization~\citep{see2017get}.
A common practice for evaluating the generation quality is to compute the similarity of the generated outputs against ground-truth references~\citep{papineni-etal-2002-bleu, lin-2004-rouge, banerjee-lavie-2005-meteor, zhang2019bertscore}. 
In open-domain dialogue areas, however, the set of potential responses for a single dialogue history is extremely large, as a conversation can evolve in many ways. Due to this very nature of dialogues, reference-based metrics show a poor correlation with human evaluations~\citep{liu-etal-2016-evaluate}.  

\begin{figure}[t!]
\centering
\includegraphics[width=\columnwidth]{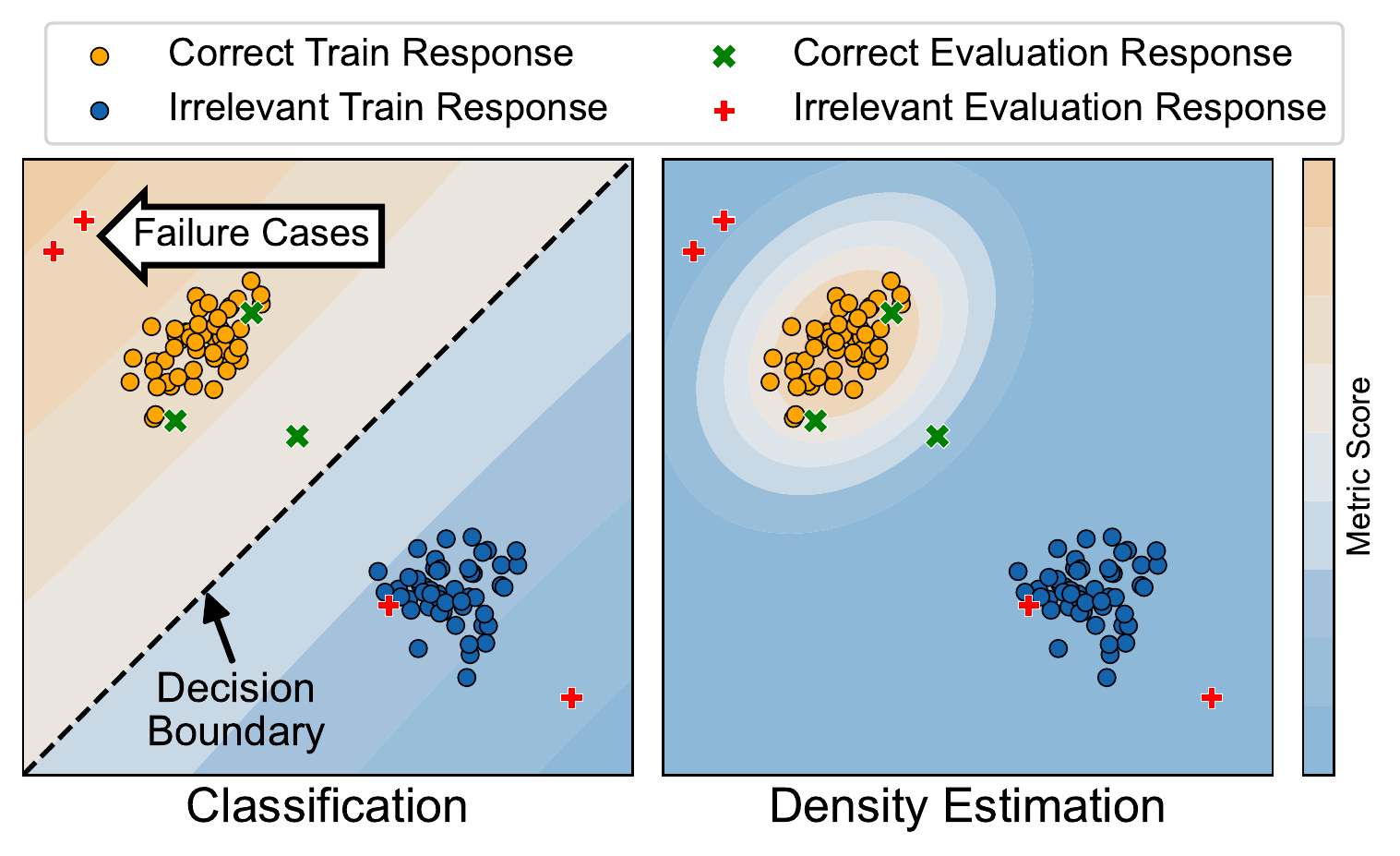}
\caption{A motivating example of \textsc{DEnsity}. 
The color bar on the right indicates the score.
Left: A classifier-based metric fails to give low scores to irrelevant responses. 
Right: A metric with density estimation successfully penalizes the irrelevant responses.}
\label{fig:toy}
\vspace{-0.5cm}
\end{figure}
To remedy this, recent studies proposed various reference-free and model-based metrics for dialogue evaluation. 
Many of them focus on estimating the relevance of a response to a dialogue history. 
For instance, \citet{tao2018ruber} propose a classification model that is trained to distinguish a correct response for a dialogue history from irrelevant responses. 
After training, the model is used to evaluate responses by predicting how likely the response would follow the dialogue history. 
These classifier-based metrics have shown a higher correlation with the human evaluations than the reference-based metrics~\cite{ghazarian-etal-2019-better, mehri-eskenazi-2020-usr, maude, zhang-etal-2021-dynaeval}. %

However, the goal of training such metrics is to find the decision boundary of classifying the training examples. 
As shown in Fig.~\ref{fig:toy}, if an irrelevant example from an unseen distribution is far from the decision boundary, but is on the same side as the correct train responses, a neural classifier will incorrectly give a high score~\citep{hendrycks2016baseline, liang2018enhancing}. 
Therefore, a metric that assumes that a generated response comes from a distribution similar to the training data is not reliable. 
Due to this misalignment of goals, classifier-based metrics may not be suitable for evaluating open-domain dialogue systems.




A similar misalignment is found in different tasks that utilize neural classifiers, such as out-of-distribution (OOD) detection.
Instead of using a prediction score from classifiers, studies utilized an alternative approach, in which the goal is to detect abnormal examples by estimating their densities on the feature space of a classifier~\citep{lee2018simple,winkens2020contrastive,zhou2021contrastiveOOD}, and showed impressive results in OOD detection. 





Inspired by the benefits of the density estimation approaches in OOD detection, we propose \textbf{\textsc{DEnsity}}, a new \textbf{D}ialogue \textbf{E}valuation metric using De\textbf{\textsc{nsity}} Estimation.
\textsc{DEnsity} measures the density of the response on the feature distribution of human conversations. Specifically, a response selection model is utilized as a feature extractor to obtain representations of both the human responses in the dialogue corpus and the system generated response. Human response features are fitted on a multivariate Gaussian distribution, and the density of the generated response on the human distribution is estimated using the Mahalanobis distance. 
Moreover, we adopt contrastive learning to further compress the features of appropriate human responses.
Looking at the right figure on Fig.~\ref{fig:toy}, unlike the classifier-based metric, density estimation properly assess the evaluation examples, and assign correct scores to relevant and irrelevant responses.  
Preliminary studies suggest that our density estimation based metric can be more robust to various failures of dialogue systems than the classifier-based metrics. 
Experiments on four turn-level response evaluation datasets show that \textsc{DEnsity} correlates better with the human evaluation than other metrics.
We summarize our contributions as follows:
\begin{enumerate}
    \item We introduce a new reference-free learnable metric for open-domain dialogue systems, which estimates the density of responses on the distribution of human conversations.
    \item Extensive experiments on multiple datasets demonstrate that, compared to other baseline metrics, \textsc{DEnsity} correlates better with human evaluations, confirming its superiority.
\end{enumerate}

\section{Related Work}
\begin{figure*}[t]
\centering
\begin{tabular}{l}
     \includegraphics[width=\textwidth]{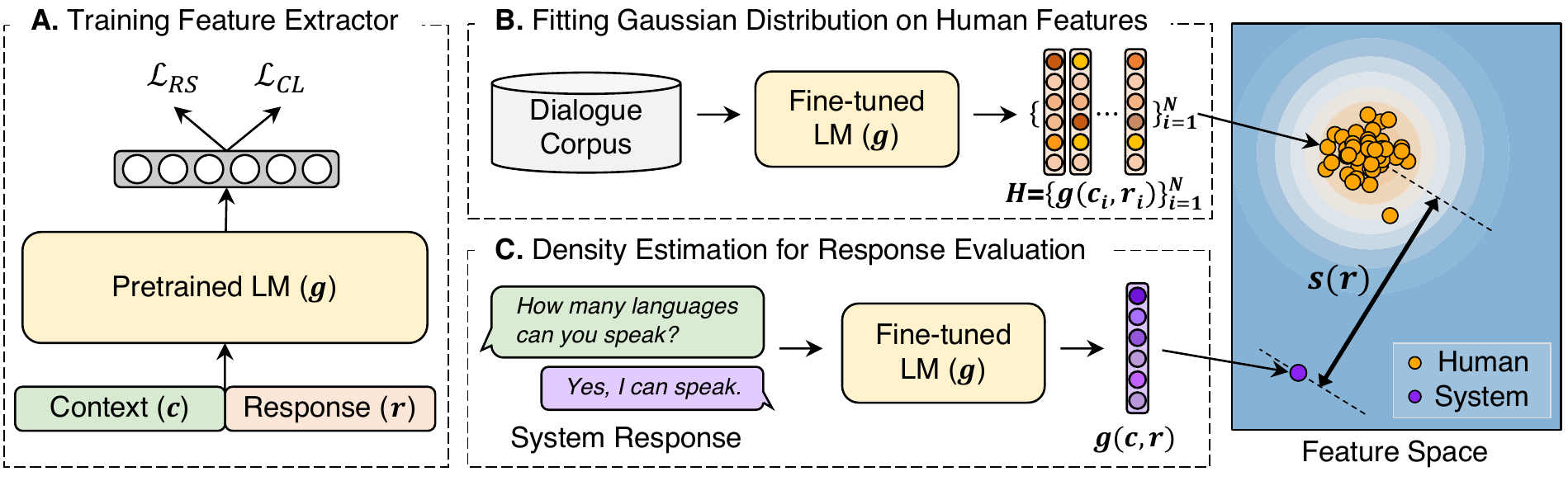}
\end{tabular}
\caption{The overall illustration of \textsc{DEnsity}.
We first train a response selection model~(\cref{sec:method1}), and employ it to extract features of both human conversations and a generated response~(\cref{sec:method2}). 
The generated response is evaluated by measuring its density on the distribution of human features~(\cref{sec:method3}).}
\label{fig:overall}
\vspace{-0.3cm}
\end{figure*}

Building a reliable automatic metric for open-domain dialogue systems is a difficult task. 
Traditional reference-based metrics, such as BLEU~\citep{papineni-etal-2002-bleu}, ROUGE~\citep{lin-2004-rouge}, or METEOR~\citep{banerjee-lavie-2005-meteor}, show a low correlation with human evaluations~\citep{liu-etal-2016-evaluate}. 
Due to the lack of a reliable automatic metric, many studies rely on human annotators to manually evaluate their systems, which can be expensive and time-consuming.
To resolve this, \citet{lowe-etal-2017-adem} propose a supervised regression model to estimate the quality of dialogue response directly. 
Despite its superior performance, the supervised regression model requires human-annotated quality scores for training, which reduces the overall generalizability of such models. 
Therefore, recent studies propose unsupervised learning-based metrics to evaluate the relevance of a response to a given dialogue history. 
For instance, \citet{tao2018ruber} train an classification model that learns to discriminate the original response from randomly sampled negative responses. 
Furthermore, researchers have extended these classifier-based metrics with various techniques. 
\citet{ghazarian-etal-2019-better} leverage pre-trained language models (LMs) to improve the evaluation performance. 
Several works aim to make hard negative samples used in training through various strategies~\citep{bak-oh-2020-speaker, maude, gupta-etal-2021-synthesizing, park-etal-2021-generating,lee-etal-2022-pneg}. 
\citet{sai-etal-2020-deb} suggest a pre-training strategy for dialogue evaluation along with a public release of an human-annotated adversarial dataset.
\citet{huang2020grade} leverage an external knowledge source~\cite{speer2017conceptnet} to augment an evaluation model. 
\citet{zhang-etal-2021-dynaeval} propose a new graph-based model to focus on the interactive and multi-turn natures of a dialogue. 
Another line of research evaluates a response by measuring the likelihood of words in a response~\citep{mehri-eskenazi-2020-usr, pang-etal-2020-towards}. 
This approach usually employs pre-trained LMs~\citep{devlin-etal-2019-bert, radford2019gpt2} to estimate the likelihood. 
Our work is distinct from previous studies in that we evaluate a response by measuring its similarity to human responses by exploiting the rich information presented in the feature space of a neural network. 

Numerous studies propose to understand and evaluate artificial responses from neural generation models. 
For instance, \citet{holtzman2019curious} report that neural language models often create incoherent and repetitive sequences. 
\citet{pillutla2021mauve} compare the distribution of a generated text against the ones written by humans in a quantized embedding space. 
In the dialogue domain, \citet{xiang2021assessing} measure the distance between the distributions of generated conversations and real-world conversations to compare at the system-level. 
Unlike the previous studies, we focus on measuring the extent to which a generated conversation is similar to real human conversations. 
Our work is inspired by previous studies that leverage the representation space of neural networks to detect out-of-distribution (OOD) or adversarially curated examples~\citep{lee2018simple,winkens2020contrastive,xu-etal-2020-deep,zhou2021contrastiveOOD}. 
Instead of detecting such abnormal instances, however, we aim to judge the quality of generated responses by considering their representations.

\section{\textsc{DEnsity}: Open-domain Dialogue Evaluation using Density Estimation}
\label{sec:method}
We present \textsc{DEnsity}, which evaluates a response by measuring its density on a distribution of human responses.
We first train a response selection model that learns to distinguish a correct response from random responses (\cref{sec:method1}). 
The selection model is employed as a feature extractor to obtain features of both human responses in a dialogue corpus and a generated response~(\cref{sec:method2}). 
We then evaluate the generated response on the distribution of human features with Gaussian discriminant analysis and Mahalanobis distance~(\cref{sec:method3}). 
We also introduce our contrastive loss to obtain better features in \cref{sec:method4}.
Fig.~\ref{fig:overall} illustrates the overall pipeline of \textsc{DEnsity}.

\subsection{Training Response Selection Model for Feature Extraction}
\label{sec:method1}
The response selection model learns to find the next utterance for a given dialogue history among the response candidates. 
The response candidates contain multiple negative responses that are incorrect and not suitable as the next utterance for the given dialogue history. 
Formally, $c$ represents a dialogue context that consists of multiple utterances between two speakers. 
The response candidate $C$ contains one answer response $r_p$ and $|C|-1$ negative responses 
$\{r_{n_i}\}_{i=1}^{|C|-1}$ 
that are randomly sampled from a dialogue corpus. 
The selection model is trained to distinguish a positive pair $(c,r_p)$ from the negative pairs $(c, r_{n_i})$ as follows: 
\begin{equation}
\mathcal{L}_{RS}=-\log\frac{e^{f(c, r_p)}}{e^{f(c,r_p)} + \sum_{i=1}^{|C|-1} e^{f(c,r_{n_i})}}
\end{equation}
where $f(\cdot,\cdot)$ denotes the prediction score of the selection model to a context-response pair. 
We implement the selection model with transformer~\citep{vaswani2017attention}-based cross-encoder architecture, where the concatenation of a dialogue history and a response $[c; r]$, along with the \texttt{[SEP]} token, is fed into a pre-trained transformer encoder $g$. 
The $d$-dimensional output representation from the transformer encoder $h=g(c,r) \in \mathbb{R}^{d}$ is then transformed into a single scalar value $f(c,r) = Wh_r$ with a linear layer $W \in \mathbb{R}^{1 \times d} $. 
In this work, we use the \texttt{[CLS]} vector from the transformer encoder as the output representation.

\subsection{Fitting Gaussian Distribution on Features}
\label{sec:method2}
After training the selection model, we encode all positive training pairs in a dialogue corpus with the encoder $g$ to obtain their output representations 
$H=\{h_i\}_{i=1}^N=\{g(c_i,r_i)\}_{i=1}^N$, 
where $N$ denotes the total number of positive pairs. 
We then train a single-class generative classifier by fitting the multivariate Gaussian distributions as follows:
\begin{equation}
\mu=\frac{1}{N}\sum_{i=1}^N h_i, 
\Sigma=\frac{1}{N}\sum_{i=1}^N (h_i-\mu)(h_i-\mu)^\mathrm{T}
\end{equation}
where $\mu$ and $\Sigma$ denote the empirical mean and covariance matrix of features, respectively. 
Note that both $\mu$ and $\Sigma$ can be calculated only once, and be used at any time for response evaluation. 

\begin{figure}[t!]

\begin{subfigure}{0.48\textwidth}
\centering
\includegraphics[width=\linewidth]{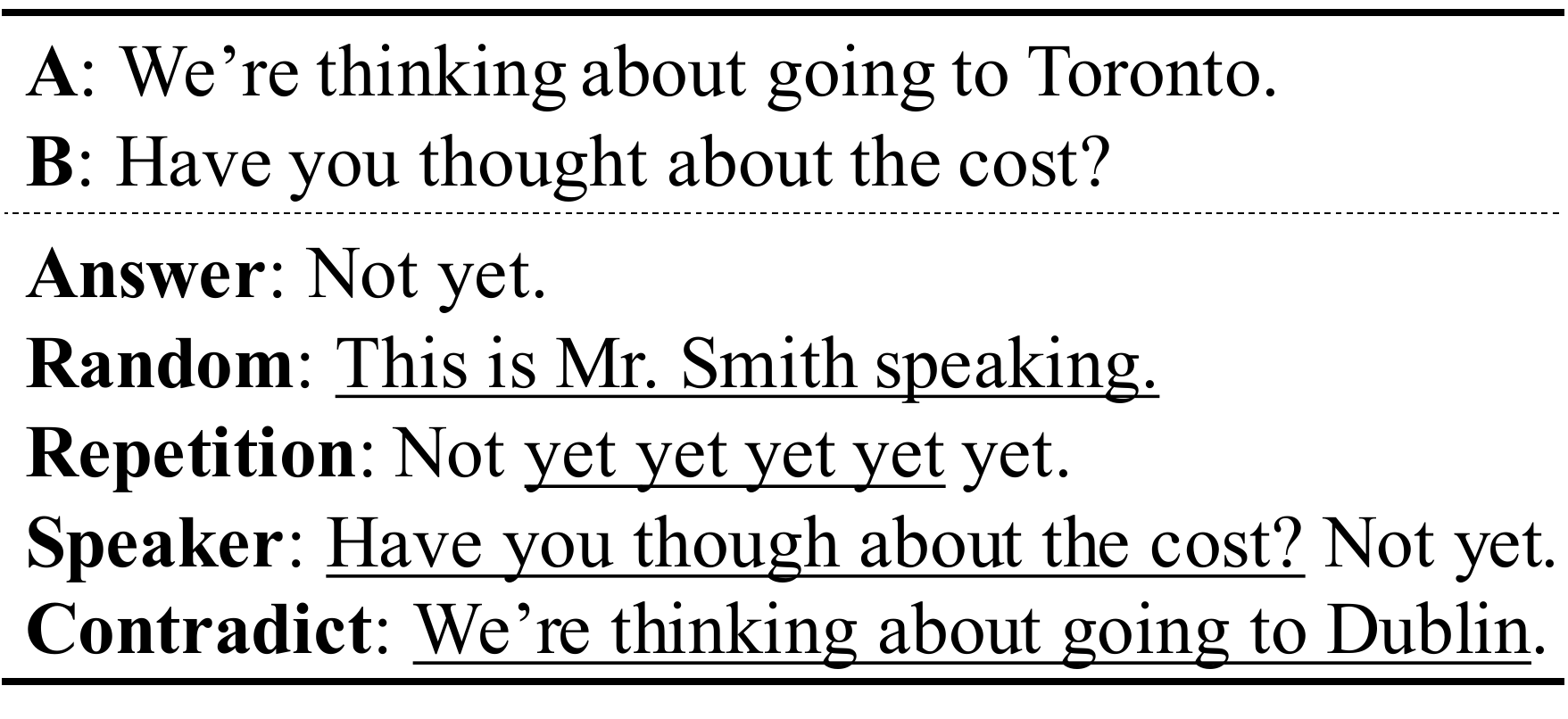}
\caption{Examples of Adversarial Responses.}
\label{fig:attack_examples}
\end{subfigure}

\begin{subfigure}{0.48\textwidth}
\includegraphics[width=\linewidth]{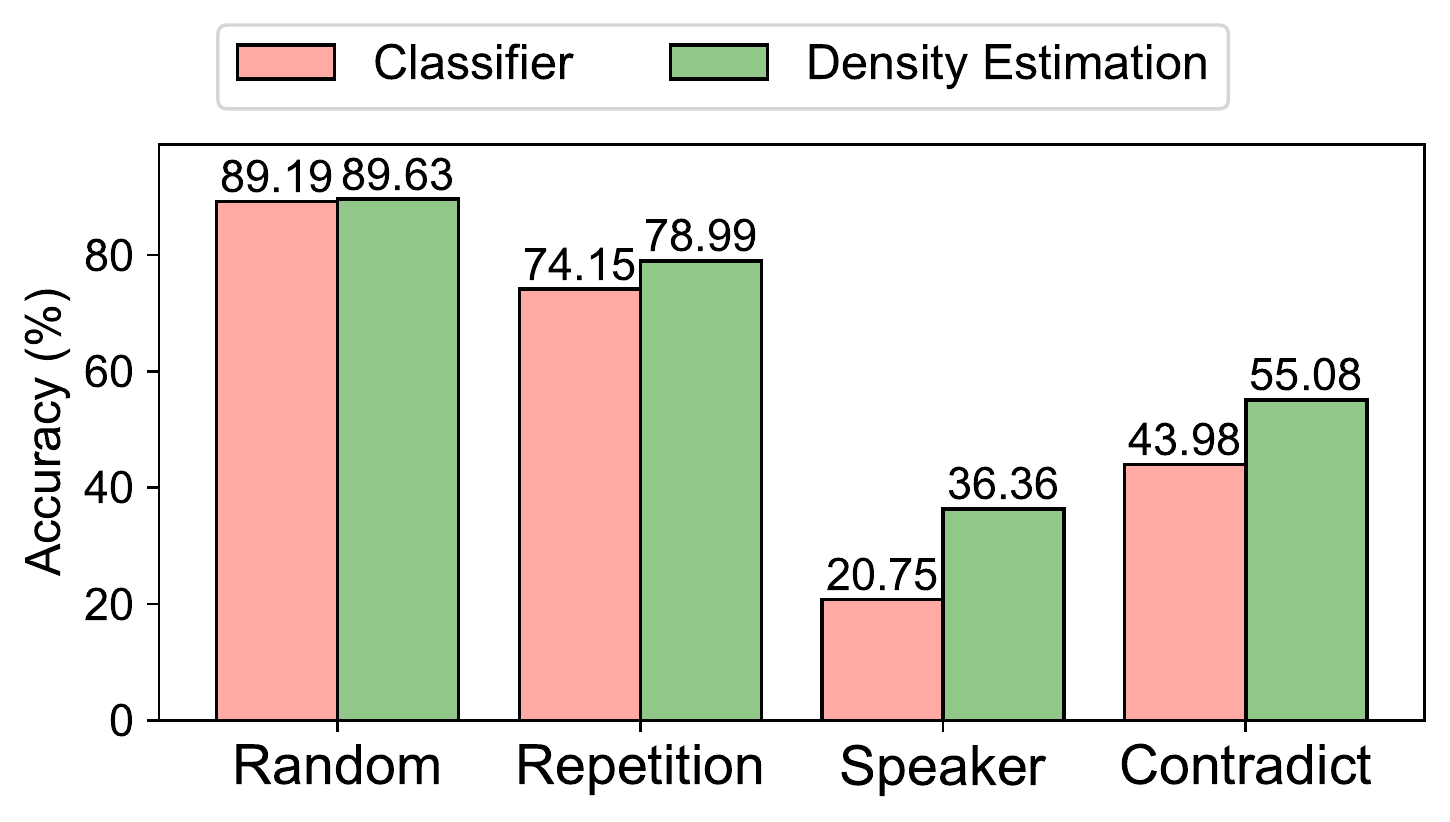}
\caption{Evaluation Accuracy}
\label{fig:comparison_sub_boxplot}
\end{subfigure}

\caption{(a)  Examples of different adversarial responses along with the original answer response. Changes to the original answer are highlighted with \underline{underline}. (b) Accuracy of metrics on different negative responses. \textit{Speaker} and \textit{Contradict} denote the Speaker-Sensitiveness and Contradiction types, respectively.}
\label{fig:adv_experiment}
\vspace{-0.3cm}
\end{figure}

\subsection{Response Evaluation with Density Estimation}
\label{sec:method3}
To evaluate a response $r$ generated by a dialogue system for a dialogue context $c$, we first obtain its feature with encoder $g$. 
We then estimate its density on the Gaussian distribution $\mathcal{N}(\mu,\Sigma)$ using a Mahalanobis distance as follows\footnote{We omit $c$ from $s(r)$ for simplicity.}:
\begin{equation}
s(r)=-(g(c, r)-\mu)\Sigma^{-1}(g(c, r)-\mu)^\mathrm{T}
\label{eq:mahalanobis_distance}
\end{equation}
where $\Sigma^{-1}$ denotes a pseudo-inverse matrix of $\Sigma$.
The distance value $s(r)$ of a response $r$ is used as the score assigned by our metric. 
In other words, we regard the distance between the response and the distribution of human responses as an indicator of its quality. 
Therefore, a high-quality response will receive a high score from our metric, while a low quality response will receive a low score.

\noindent \textbf{Does a density estimation based metric actually work?} Note that many previous studies leverage the prediction of a classifier that is trained to distinguish a correct response from others~\citep{tao2018ruber,mehri-eskenazi-2020-usr,maude}, which is similar to our selection model’s prediction score $f(c, r)$.
In contrast, our metric only uses the intermediate output of the model to estimate the density of a generated response. 
In our preliminary study, we compare the classifier-based approaches with density estimation by observing their behaviors on adversarially manipulated responses.
Specifically, we use a dataset released by \citet{khalid-lee-2022-explaining} that is designed to probe the robustness of dialogue evaluation metrics. 
The dataset consists of three components: (1) dialogue history, (2) answer response, and (3) adversarial response. 
The adversarial response is created by manipulating the original answer response with various strategies. 
An evaluation metric should assign a higher score to the answer response than the adversarial response. 
We use the following three adversarial 
types, which reflect errors that frequently occur in dialogue systems: (1) \textit{Repetition}: repeats words in the answer, (2) \textit{Speaker-Sensitiveness}: concatenates the last utterance from the dialogue history to the answer, (3) \textit{Contradiction}: corrupts the contextual information in the answer. 
We also compute the accuracy on a random negative example to check whether the density estimation based metric can perform well in the original training task. 
Fig.~\ref{fig:attack_examples} presents sample responses of different attack types. 

From the results in Fig.~\ref{fig:comparison_sub_boxplot}, we first observe that our density estimation based metric can distinguish random responses on par with a classifier-based metric. 
In terms of accuracy on the adversarial responses, the density estimation based metric performs better than the classifier-based metric. 
Previous literature on OOD detection task~\citep{lee2018simple, xu-etal-2020-deep} reports similar trends to our observation, where the softmax-based neural classifiers are prone to make overly confident predictions~\cite{hendrycks2016baseline, liang2018enhancing} on abnormal instances.
These results imply that the feature space derived by $g$ contains sufficient information to perform the original task, and the density estimation based metric can be more effective than the classifier-based metric in detecting various failures made by dialogue systems. 

\subsection{Enhanced Feature Space with Contrastive Learning}
\label{sec:method4} 
Our feature extractor $g$ is trained to make features that can discriminate the correct response from the incorrect ones for the same dialogue history. 
Therefore, features of a positive pair from a different dialogue history may not be easily distinguished from features of negative responses, which can reduce the performance of our metric.
To resolve this, 
we adopt a supervised contrastive loss~\citep{scl,gunel2020supervised,zhou2021contrastiveOOD}, which regards all positive pairs in the training set as the same class. 
The loss function encourages the representations of positive pairs to be closer, while increasing the discrepancy from the representations of negative pairs. 
Formally, given a batch of $|B|$ dialogue history, the training objective for contrastive learning is:
\begin{equation}
\label{eq:scl}
\mathcal{L}_{CL}=\sum_{i=1}^{|B|}\frac{-1}{P(i)}\sum_{p \in P(i)} \log \frac{e^{z_{p_i} \cdot z_{p}/\tau} }{\sum\limits_{a \in B(i)} e^{z_{p_i} \cdot z_a / \tau}}
\end{equation}
where $z$ denotes a L2-normalized \texttt{[CLS]} vector of a context-response pair from the transformer encoder $g$, and $p_i$ denotes the $i$th positive pair in the batch.
$P(i)$ denotes all positive context-response pairs in the batch except for $p_i$, and $B(i)$ denotes all context-response pairs in the batch except for $p_i$.\footnote{$|P(i)|$ and $|B(i)|$ are $|B|-1$ and $|B|\times|C|-1$.}
$\tau>0$ is a temperature scaling hyperparameter.
The final training objective of selection model is 
\begin{equation}
\label{eq:total_loss}
\mathcal{L}_{total}=\mathcal{L}_{RS} + \lambda \mathcal{L}_{CL}
\end{equation}
where $\lambda>0$ is a hyperparameter.

\section{Experiments}
\subsection{Dataset}
We conduct experiments on two representative open-domain dialogue datasets.

\noindent \textbf{DailyDialog}~\citep{li-etal-2017-dailydialog} is a multi-turn dialogue dataset written by human annotators. The dataset is designed to cover various topics and relationships in our daily life. The dataset consists of 13,118 multi-turn conversations.
\noindent \textbf{ConvAI2}~\citep{dinan2020second} is a crowd-sourced dataset, where two speakers continue a conversation with their assigned personal information. The dataset consists of 17,878 multi-turn conversations. 

We use four turn-level dialogue evaluation datasets to compare the performance of different evaluation metrics. In the dataset, each example consists of a dialogue history, an answer response, a generated response by a dialogue system, and a quality score judged by human annotators. 
The details of evaluation datasets are as follows:

\noindent \textbf{DailyDialog-Zhao}~\citep{zhao2020designing} contains 900 evaluation examples from six different dialogue systems. 
The dataset is derived from the DailyDialog dataset, which is used as a training corpus of dialogue systems and a source of dialogue histories. 
The “overall” score is used in our experiments.
\noindent \textbf{ConvAI2-USR}~\citep{mehri-eskenazi-2020-usr} is derived from the ConvAI2 dataset and contains 240 evaluation examples from three dialogue systems.
The “Overall Quality” score is used in our experiments. 
\noindent \textbf{DailyDialog-GRADE} and \textbf{ConvAI2-GRADE} are datasets released by \citet{huang2020grade}, each of which is based on the DailyDialog and ConvAI2 datasets.
DailyDialog-GRADE and ConvAI2-GRADE contain 300 examples from two systems and 600 examples from four systems, respectively.

\begin{table*}[t!]
\tiny
\centering
\begin{adjustbox}{width=1.0\textwidth}
\begin{tabular}{
    lcccccccc
}
\toprule
                   & \multicolumn{2}{c}{\textbf{\begin{tabular}[c]{@{}c@{}}DailyDialog\\Zhao\end{tabular}}} 
                   & \multicolumn{2}{c}{\textbf{\begin{tabular}[c]{@{}c@{}}ConvAI2\\USR\end{tabular}}} 
                   & \multicolumn{2}{c}{\textbf{\begin{tabular}[c]{@{}c@{}}DailyDialog\\GRADE\end{tabular}}} 
                   & \multicolumn{2}{c}{\textbf{\begin{tabular}[c]{@{}c@{}}ConvAI2\\GRADE\end{tabular}}} \\
                   
\textbf{Model}     & \multicolumn{1}{c}{$r$} & \multicolumn{1}{c}{$\rho$}  & \multicolumn{1}{c}{$r$} & \multicolumn{1}{c}{$\rho$} & \multicolumn{1}{c}{$r$} & \multicolumn{1}{c}{$\rho$} & \multicolumn{1}{c}{$r$} & \multicolumn{1}{c}{$\rho$} \\ 
\cline{0-8}\noalign{\vskip 0.3ex}
\multicolumn{9}{c}{\textit{Reference-based}} \\ 
\cline{0-8}\noalign{\vskip 0.75ex}
BLEU             & 8.02$^{*}$ & 2.77$^{*}$ & 11.22$^{*}$ & 12.43$^{*}$ & 14.15$^{*}$ & 10.70$^{*}$ & 10.69$^{*}$ & 12.36 \\
ROUGE              & 14.96 & 9.79$^{*}$ & 10.96$^{*}$ & 9.64$^{*}$  & 10.98$^{*}$ & 3.12$^{*}$  & 11.82  & 11.56 \\
METEOR             & 8.99$^{*}$ & 5.36$^{*}$ & 17.99$^{*}$ & 18.80  & 11.94$^{*}$ & 7.54$^{*}$  & 22.48  & 22.50 \\
EmbAvg.             & 7.32$^{*}$ & 7.93$^{*}$ & 11.82$^{*}$ & 14.51$^{*}$ & 2.44$^{*}$  & 4.08$^{*}$  & 19.52  & 21.47 \\
EmbExt.             & 18.83 & 17.81 & 15.70$^{*}$ & 14.27$^{*}$ & 13.56$^{*}$ & 9.77$^{*}$  & 15.94  & 16.82 \\
EmbGrd.             & 15.47 & 15.35 & 9.16$^{*}$  & 6.51$^{*}$  & 11.06$^{*}$ & 8.29$^{*}$  & 21.54  & 22.14 \\
BERTScore          & 15.32 & 15.36 & 15.16$^{*}$ & 12.27$^{*}$ & 12.88$^{*}$ & 9.94$^{*}$  & 22.48  & 22.50 \\
SimCSE             & 24.07 & 22.00 & 28.83  & 29.31  & 21.68  & 18.22  & 23.69  & 24.28 \\
BLEURT             & 14.42 & 12.88 & 16.09$^{*}$ & 15.25$^{*}$ & 17.45  & 12.24$^{*}$ & 10.16$^{*}$ & 10.68$^{*}$ \\ 
\cline{0-8}\noalign{\vskip 0.3ex}
\multicolumn{9}{c}{\textit{Perplexity-based}} \\ \hline
\cline{0-8}\noalign{\vskip 0.75ex}
USR-MLM            & 38.39 & 39.84 & 7.01$^{*}$  & 13.32$^{*}$ & 18.32  & 20.60  & 14.38  & 9.48  \\
GPT2-Coh.           & 44.19 & 45.46 & 20.52  & 19.77  & 23.40  & 25.83  & 43.34  & 43.99 \\ 
FED                & -7.30 $^{*}$ & -6.58$^{*}$ & -2.03$^{*}$ & 0.58$^{*}$ & 2.56$^{*}$ & 0.13$^{*}$  & -14.29 & -14.55 \\ 
\cline{0-8}\noalign{\vskip 0.3ex}
\multicolumn{9}{c}{\textit{Classifier-based}} \\  
\cline{0-8}\noalign{\vskip 0.75ex}
BERT-RUBER         & 36.07 & 35.52 & 14.61$^{*}$ & 17.05$^{*}$ & \uline{28.29} & \uline{25.99} & 5.73$^{*}$ & 2.65$^{*}$ \\
USR-Retrieval       & \uline{48.77} & \uline{51.61} & 49.96 & \uline{59.65} & 27.47 & 23.84 & 40.30 & 39.98 \\
USL-H              & 37.19 & 38.62 & \uline{52.36}  & 53.36& 10.90$^{*}$ & 9.72$^{*}$ & \uline{44.89} & \uline{45.47} \\
FlowScore           & 11.97 & 12.64 & -9.06$^{*}$ & -7.49$^{*}$& 13.00$^{*}$  & 14.78$^{*}$ & 5.92$^{*}$ & 6.81$^{*}$ \\
DynaEval           & 27.21 & 27.16 & 10.08$^{*}$ & 10.67$^{*}$& 14.96$^{*}$  & 15.89$^{*}$ & 23.89 & 22.83 \\ 
\cline{0-8}\noalign{\vskip 0.3ex}
\multicolumn{9}{c}{\textit{Density Estimation based~(Ours)}} \\ 
\hline
\cline{0-8}\noalign{\vskip 0.75ex}
\textsc{DEnsity}     & \textbf{56.81} & \textbf{57.03} & \textbf{57.03}  & \textbf{62.97} & \textbf{30.33}  & \textbf{29.45} & \textbf{48.01} & \textbf{48.62} \\
\bottomrule
\end{tabular}
\end{adjustbox}
\caption{The correlations between automatic metrics and human evaluation on four evaluation datasets. $r$ is Pearson correlation, and $\rho$ is Spearman's rank correlation coefficient. The highest and the second highest scores in each column are highlighted in \textbf{bold} and \underline{underline}, respectively. \textit{GPT2-Coh.} denotes the GPT2-Coherency metric. All values with p > 0.01 are marked with *.}
\label{tab:main_table}
\vspace{-0.3cm}
\end{table*}

\subsection{Baselines}
We compare our method against the following baseline metrics.
Note that the training dataset of reference-free metrics is the same as the original dialogue dataset from which the evaluation dataset is derived.
Further implementation details of the baseline metrics are available in Appendix~\ref{appendix:baseline}.

\noindent \textbf{BLEU}, \textbf{ROUGE}, and \textbf{METEOR} measure word overlap of hypothesis against references.

\noindent \textbf{Embedding Average/Greedy/Extrema}~\citep{liu-etal-2016-evaluate} compute the similarity between an answer and generated responses with a distributed word representation. 

\noindent \textbf{BERTScore}~\citep{zhang2019bertscore} use a pre-trained LM to obtain the contextualized embedding of responses for similarity comparison.

\noindent \textbf{SimCSE}~\citep{gao-etal-2021-simcse} adopts a self-supervised contrastive learning for better sentence embeddings. We calculate the cosine similarity between the embeddings of an answer and generated responses.

\noindent \textbf{BLEURT}~\citep{sellam-etal-2020-bleurt} is a reference-based metric pretrained on a synthetic dataset for an evaluation of machine translation systems.

\noindent \textbf{USR-MLM}~\citep{mehri-eskenazi-2020-usr} replaces each token in a response to \texttt{[MASK]}, and aggregates the likelihood of each token when they are conditioned on the context by using masked language modeling by BERT~\citep{devlin-etal-2019-bert}.

\noindent \textbf{GPT2-Coherency}~\citep{pang-etal-2020-towards} measures the perplexity of a response conditioned on its dialogue history by using GPT-2~\citep{radford2019gpt2}.

\noindent \textbf{FED}~\citep{mehri2020fed} evaluates a response by measuring the perplexity of follow-up utterances designed by the authors.

\noindent \textbf{BERT-RUBER}~\citep{ghazarian-etal-2019-better} replaces the word embedding layer in the original RUBER~\citep{tao2018ruber} with BERT. 

\noindent \textbf{USR-Retrieval}~\citep{mehri-eskenazi-2020-usr} learns to distinguish the next utterance of a given dialog history from a random response. After training, the score for which the response is predicted to be the next utterance is used for evaluation.

\noindent \textbf{USL-H}~\citep{phy2020deconstruct} combines the predictions from multiple models trained on different tasks for configurable response evaluation.

\noindent \textbf{FlowScore}~\citep{flowscore} compares a response's semantic influences and those expected by a response generation model. 

\noindent \textbf{DynaEval}~\citep{zhang-etal-2021-dynaeval} adopts a graph-based architecture to reflect the interaction between speakers in a dialogue, and is trained to distinguish the original dialogue from the corrupted ones.


\subsection{Implementation Details}
We use BERT~\citep{devlin-etal-2019-bert} released by \citet{wolf-etal-2020-transformers} to initialize our response selection model.\footnote{\url{bert-base-uncased} is used.}
We set $\tau$ and $\lambda$ in Eq.~\ref{eq:scl} and Eq.~\ref{eq:total_loss} to 0.1 and 1.0, respectively.
The selection model is trained for 10 epochs, and the  initial learning rate is set to 5e-5. 
AdamW~\citep{adamw} optimizer is used for optimization, and the linear learning rate scheduler is used with 1000 warm up steps. 
The maximum number of tokens in the input sequence is set to 256, and the batch size is set to 16.
The number of negative responses for a dialogue history is set to 15 for DailyDialog dataset, and 19 for ConvAI2 dataset. 
We use the square root value of the distance in Eq.~\ref{eq:mahalanobis_distance}, since it empirically shows a better performance. 
Further implementation details are in Appendix~\ref{appendix:implementation}.

\begin{figure*}[t!]
\centering
\begin{tabular}{l}
     \includegraphics[width=.9\textwidth]{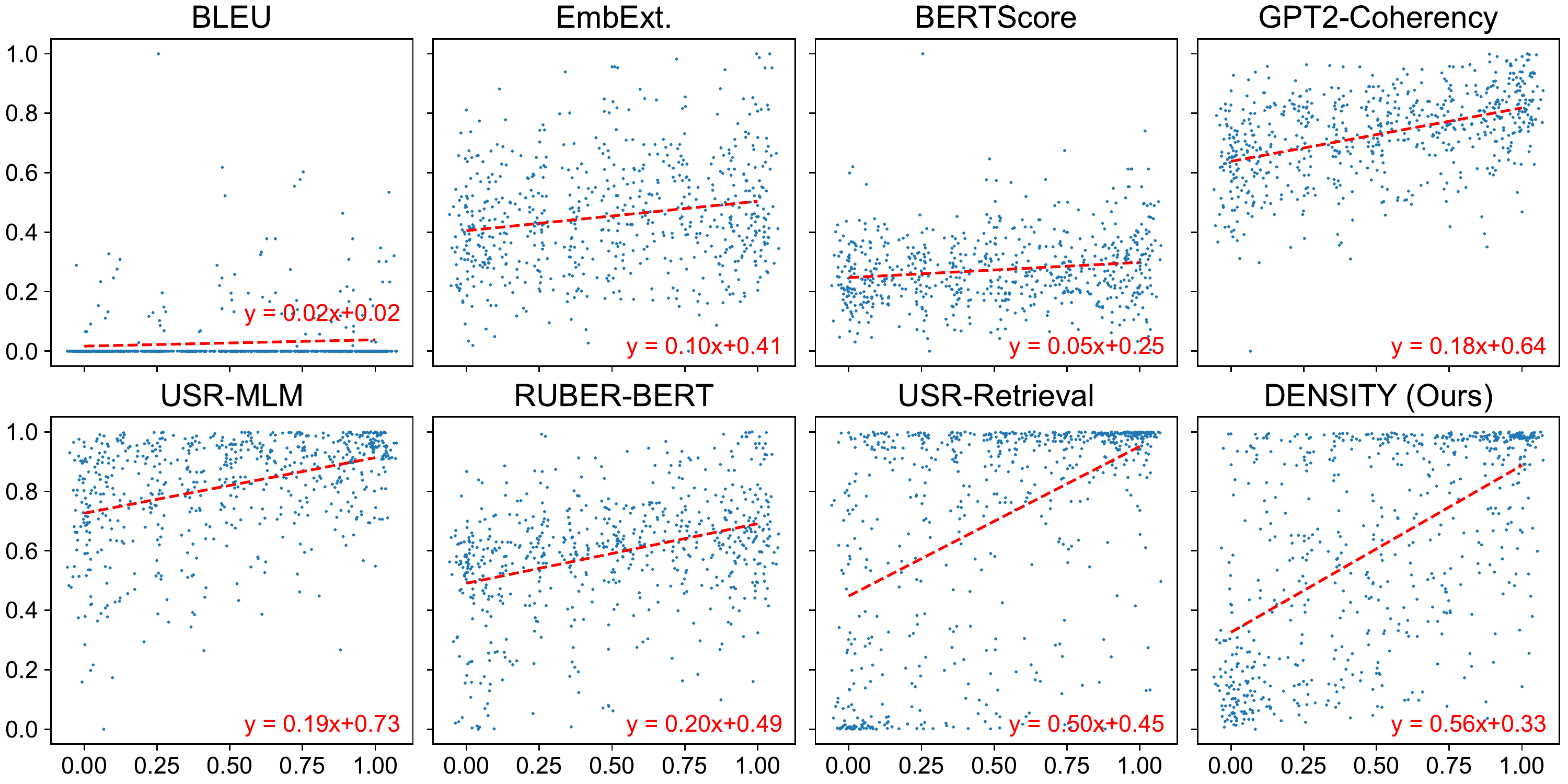}
\end{tabular}
\caption{Scatter plots between human scores and model predictions on DailyDialog-Zhao dataset. 
Each point indicates a response, and the x and y values of each point indicate human score and metric score, respectively. 
Both x and y values are normalized into [0,1] scale.
For a better visualization, we add a Gaussian noise sampled from $\mathcal{N}(0, 0.03^2)$ to human scores~\citep{lowe-etal-2017-adem,bak-oh-2020-speaker}. The red line indicates a linear regression.}
\label{fig:scatter}
\vspace{-0.3cm}
\end{figure*}

\section{Results}
\label{sec:results}

\subsection{Main Results}
\label{sec:main_results}

Table~\ref{tab:main_table} shows the results of the response evaluation task on each dataset. 
Pearson correlation coefficient~($r$) and Spearman’s rank correlation coefficient~($\rho$) are used to compare the model predictions with human scores. 
We first observe that \textsc{DEnsity} achieves the highest correlation with human scores in all evaluation datasets. 
In terms of baseline metrics, n-gram overlap-based metrics like BLEU usually show a low correlation with human scores.
These observations are consistent with previous studies~\citep{liu-etal-2016-evaluate}, where the metrics relying on a comparison with the answer response are not appropriate in the dialogue domain. 
Reference-based metrics like SimCSE and BLEURT, which do not rely on the n-gram overlap, show an improved performance, but their correlations are still relatively lower than the reference-free metrics.
The reference-free metrics that either measure the likelihood of words in a response (e.g., USR-MLM and GPT2-Coherency) or use the classification model (e.g., BERT-RUBER, USR-Retrieval) usually show a better performance than the reference-based metrics.
Notably, classifier-based metrics like BERT-RUBER, USR-Retrieval, and USL-H often show competitive performance with \textsc{DEnsity}.
DynaEval, a well-known metric for dialogue-level evaluation, usually performs worse than other reference-free metrics, which can be attributed to the different characteristics of turn-level and dialogue-level evaluations.
Based on these results, we can confirm the validity and the effectiveness of \textsc{DEnsity}.

\subsection{Correlation Visualization}
Scatter plots in Fig.~\ref{fig:scatter} present human scores and the prediction scores of each metric. 
Each point indicates an evaluated response, and the x-axis and y-axis values indicate the human score and metric score, respectively. 
BLEU, a metric that relies on word overlap similarity, usually gives low scores close to zero, which makes it hard to be adopted as a reliable metric. 
Embedding-based metrics like Embedding Extrema and BERTScore tend to give similar scores to responses of different human scores, which fails to discriminate between high-quality responses. 
Several reference-free metrics like RUBER-BERT and GPT2-Coherency make predictions that positively correlate with human scores. 
USR-Retrieval successfully gives high scores to responses with high human scores, but it also frequently makes overly confident predictions to responses with low human scores. 
This \textit{false-positive} problem is relatively alleviated in \textsc{DEnsity}, as the number of high predictions for low human scores is decreased. 
One possible limitation of our metric is that it makes confident predictions when responses receive a score higher than 0.8 from human annotators, which implies that our model might not be calibrated well.
This behavior would be undesirable to ones that hope to find the best response among reasonably well-written responses. 
Building a well-calibrated metric for general purposes is left as future work.

\begin{table}[t!]
\small
\begin{adjustbox}{width=\columnwidth}
\begin{tabular}{lcrrrr}
\toprule
                 &  \multicolumn{1}{l}{}           & \multicolumn{2}{c}{\textbf{\begin{tabular}[c]{@{}c@{}}DailyDialog\\ Zhao\end{tabular}}} & \multicolumn{2}{c}{\textbf{\begin{tabular}[c]{@{}c@{}}DailyDialog\\ GRADE\end{tabular}}} \\
\textbf{Scoring} & \multicolumn{1}{c}{\textbf{$\mathcal{L}_{CL}$}} & \multicolumn{1}{c}{$r$}                       & \multicolumn{1}{c}{$\rho$}                       & \multicolumn{1}{c}{$r$}                        & \multicolumn{1}{c}{$\rho$}                       \\ 
\cline{0-5}\noalign{\vskip 1.0ex}
Maha.             & \checkmark                       & \textbf{56.81}                              & \textbf{57.03}                              & \textbf{30.33}                               & 29.45                                       \\
Euclidean        & \checkmark                        & 54.57                                       & 56.84                                       & 28.97                                        & 31.92                              \\
Classifier       & \checkmark                         & 53.36                                       & 55.14                                       & 28.25                                        & 29.10                                       \\
Maha.             &                                  & 46.79                                       & 51.97                                       & 24.52                                        & 27.40                                       \\
Euclidean        &                                  & 52.88                                       & 52.15                                       & 30.22                                        & \textbf{32.04}                                       \\
Classifier       &                                  & 53.17                                       & 53.66                                       & 22.58                                        & 22.12                                      \\ \bottomrule
\end{tabular}
\end{adjustbox}
\caption{Results of ablation study. $\mathcal{L}_{CL}$ indicates a contrastive loss in Eq.~\ref{eq:scl}. \textit{Maha.} and \textit{Euclidean} denote density estimation based metrics with Mahalanobis and Euclidean distance functions, respectively. 
\textit{Classifier} uses a selection model's prediction score $f(c,r)$. 
The highest score in each column is highlighted in \textbf{bold}.}
\label{tab:ablation_studys}
\vspace{-0.3cm}
\end{table}
\begin{figure}[t!]
\centering
\includegraphics[width=0.48\textwidth]{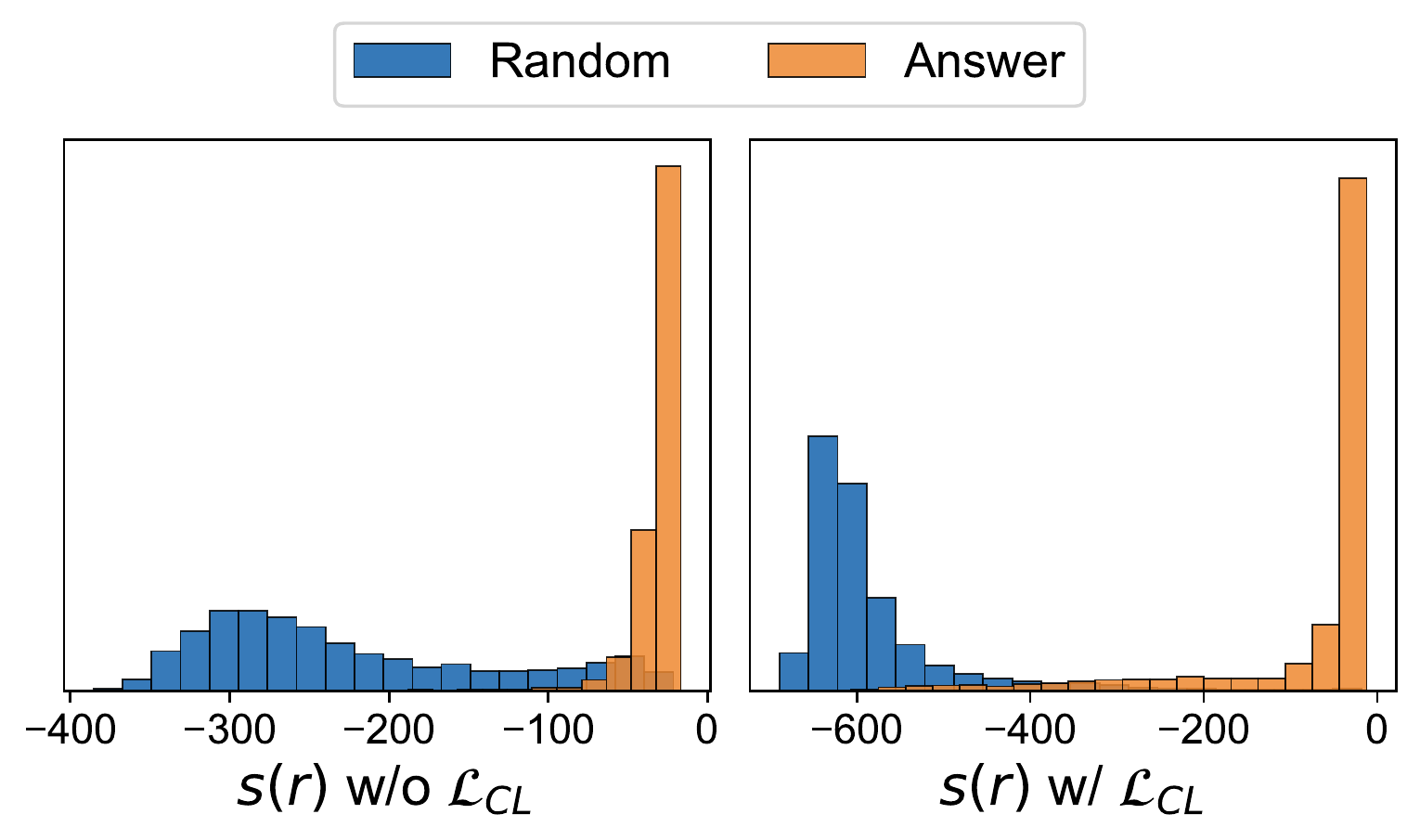}
\caption{Histogram of $s(r)$ to different response types with and without contrastive learning. \textit{Random} and \textit{Answer} denote random and answer responses, respectively.}
\label{fig:scl}
\vspace{-0.3cm}
\end{figure}

\subsection{Ablation Study}
\label{sec:ablation_study}
We conduct an ablation study to investigate the impact of each component in our metric. 
Results are shown in Table~\ref{tab:ablation_studys}. 
We first observe that the correlations of distance-based scoring functions are largely increased with contrastive learning. 
Such gain is considerably larger for Mahalanobis distance, while there are cases in which the correlation dropped with Euclidean distance. 
Regarding the comparison between different approaches for response evaluation (classifier vs density estimation), the classifier-based metric performs better than density estimation based metrics when the contrastive loss is not applied. 
These results imply that contrastive learning is indeed helpful in improving the performance of our metric, by enhancing the quality of the feature space from the selection model. 
In terms of the two different distance functions, the Mahalanobis distance with $\mathcal{L}_{CL}$ usually shows a higher correlation than the Euclidean distance.

To further investigate the impact of contrastive learning on a feature space from the selection model, we compute \textsc{DEnsity} score $s(r)$ of answer responses and random negative responses on the validation set of the original DailyDialog corpus. 
The score distributions of the selection models trained with or without $\mathcal{L}_{CL}$ objective are compared. 
The score distributions in Fig.~\ref{fig:scl} confirm that contrastive learning encourages the model to make a more discriminative feature space.

\newcommand*{\talign}[2]{%
  #1 & : #2 \\
}

\setlength\heavyrulewidth{0.25ex}
\begin{table}[t!]
\small
\centering
\resizebox{\columnwidth}{!}{%
\begin{tabular}{llll}
\toprule
\multicolumn{4}{l}{\begin{tabular}[c]{@{}l@{}}\textbf{A}: Do you want any meat today, Mrs. bird?\\ \textbf{B}: Yes, please.\\ \textbf{A}: Do you want beef or lamb?\\ 
\textbf{B (Answer)}: Beef, please.\\ 
\textbf{B (System)}: Sticks, please.\end{tabular}} \\ 

\noalign{\vskip 0.5ex}\cdashline{0-3}\noalign{\vskip 0.75ex}

BLEU& 1.00 (1.00) & USL-H& 0.60 (0.49) \\
USR-R& 0.96 (0.61) & GPT2-Coh.& 0.70 (0.37) \\
Human& 0.25 (0.26) & \textsc{DEnsity}&\textbf{0.39 (0.32)} \\ 

\midrule
\multicolumn{4}{l}{\begin{tabular}[c]{@{}l@{}}\textbf{A}: I think it's running late.\\ \textbf{B}: It should've been here 30 minutes ago.\\ \textbf{A}: It should be coming soon.\\ \textbf{B}: It better, because I'm already late for work.\\ \textbf{A}: I can't stand riding the bus.\\ 
\textbf{B (Answer)}: Me too, it's too unreliable.\\ 
\textbf{B (System)}: Shall I take a bus to get there?\end{tabular}} \\ 

\noalign{\vskip 0.5ex}\cdashline{0-3}\noalign{\vskip 0.75ex}

BLEU &  0.00 (0.00) & USL-H & 0.47 (0.34) \\
USR-R & 0.97 (0.63) & GPT2-Coh. & 0.86 (0.83) \\
Human & 0.08 (0.14) & \textsc{DEnsity}& \textbf{0.11 (0.12)}  \\ 

\midrule

\multicolumn{4}{l}{\begin{tabular}[c]{@{}l@{}}\textbf{A}: I need to go to the bank.\\ \textbf{B (Answer)}: But they are closed today.\\ \textbf{B (System)}: Okay. I'll take you there.\end{tabular}} \\ 

\noalign{\vskip 0.5ex}\cdashline{0-3}\noalign{\vskip 0.75ex}

BLEU    & 0.00 (0.59) & USL-H   & 0.63 (0.59) \\
USR-R   & 0.27 (0.21) & GPT2-Coh.   & 0.68 (0.35) \\
Human   & 1.00 (0.86) & \textsc{DEnsity}& \textbf{0.95 (0.73)} \\
\midrule
\multicolumn{4}{l}{\begin{tabular}[c]{@{}l@{}}\textbf{A}:  I did learn that.\\ \textbf{B}: So, then why did you speed up?\\ \textbf{A}: I don't know what to tell you.\\ \textbf{B}: I'm going to have to write you a ticket.\\ \textbf{A}: I understand.\\ \textbf{B (Answer)}: Here you go. Don’t do that again.\\ \textbf{B (System)}: And you'll have to pay a fine.\end{tabular}} \\ 
\noalign{\vskip 0.5ex}\cdashline{0-3}\noalign{\vskip 0.75ex}

BLEU & 0.00 (0.72) & USL-H & 0.65 (0.72) \\
USR-R& 0.65 (0.32) & GPT2-Coh.& 0.80 (0.66) \\
Human & 1.00 (0.86) & \textsc{DEnsity}& \textbf{0.98 (0.82)}  \\ 
\bottomrule
\end{tabular}%
}

\caption{Sample results of selected metrics on the DailyDialog-Zhao dataset. The score next to the metric name is a metric score normalized into the [0,1] scale, and the score in parentheses is the rank score. The rank score is the rank of the metric score divided by the total number of examples in the dataset. The closest score with the human evaluations is marked with \textbf{bold}.} 
\label{tab:case-study}
\vspace{-0.3cm}
\end{table}

\subsection{Case Study}
We present sample evaluation results of few selected metrics on the DailyDialogue-Zhao dataset in Table~\ref{tab:case-study}. While the system responses in the first and second examples look okay, taking a closer look at the dialog history, they are both incoherent. In the first example, the question asked gave two options, but the system responded with an option that was not in the original question. In the second example, the system asks if he (or she) should take the bus, even though he is already waiting for the bus. 
The low scores from human annotators highlight this incoherency. 
Baseline metrics like USR-Retrieval and GPT2-Coherency often give high scores to such responses. In contrast, \textsc{DEnsity} outputs low scores, similar to human scores. In the third and fourth example, the system responses receive high scores from the human annotators. BLEU gives low scores to both responses, as there are little word overlap between the answers and system responses. In both examples, \textsc{DEnsity} gives relatively similar scores to human scores.

\begin{table}[t!]
\centering
\tiny
\begin{adjustbox}{width=0.8\columnwidth}
\begin{tabular}{ll}
\midrule
\multicolumn{1}{c}{\textbf{Model}} & $\rho$ \\
\cline{0-1}\noalign{\vskip 0.35ex}
\multicolumn{2}{c}{\textit{Turn-level Metrics}} \\
\cline{0-1}\noalign{\vskip 0.75ex}
USR-Retrieval & 39.4 \\
GPT2-Coherency & 30.4 \\
\textsc{DEnsity} & \underline{43.3}  \\
\cline{0-1}\noalign{\vskip 0.3ex}
\multicolumn{2}{c}{\textit{Dialogue-level Metrics}} \\
\cline{0-1}\noalign{\vskip 0.75ex}
FED& 40.1$^{+}$                     \\
DynaEval      & \textbf{49.2}$^{+}$                     \\
\cline{0-1}\noalign{\vskip 0.3ex}
\multicolumn{2}{c}{\textit{Human Performance}} \\
\cline{0-1}\noalign{\vskip 0.75ex}
Human~\citep{zhang-etal-2021-dynaeval} & 83.0$^{+}$ \\
\midrule

\end{tabular}
\end{adjustbox}
\caption{The correlation between automatic metrics and human evaluation on FED dialogue-level evaluation dataset. Human performance are cited from \citet{zhang-etal-2021-dynaeval}. Scores marked with $+$ are from the original paper. The highest and second highest metric scores are highlighted in \textbf{bold} and \underline{underline}, respectively.}
\label{tab:dialogue_level}
\end{table}

\subsection{Experiments on Dialogue-level Evaluation}
\label{appendix:dialogue_level}
While our experiments generally focus on turn-level response evaluation, we also conduct experiments on dialogue-level evaluation to probe the extensibility of \textsc{DEnsity} on such tasks. 
To this end, we use the FED dialogue-level evaluation dataset~\citep{mehri2020fed}, and the “Overall” score is used to calculate the Spearman correlation. 
We compare \textsc{DEnsity} against some turn-level metrics that were competitive with \textsc{DEnsity} in our turn-level evaluation (USR-Rtv. and GPT2-Coh.).
To extend the turn-level evaluation metrics to a dialogue-level evaluation, we simply evaluate every turn in a dialogue, and average their scores. 
We also include the results of the FED~\citep{mehri2020fed} and DynaEval~\citep{zhang-etal-2021-dynaeval} models from the original papers.\footnote{We use the results of FED model with 345M parameters, which has a comparable size with other models.} The results are shown in Table~\ref{tab:dialogue_level}.

The results show that \textsc{DEnsity} shows a higher correlation than other turn-level metrics. This result shows that although our current metric is not explicitly designed to handle dialogue-level evaluations, there is potential for utilizing a density estimation-based evaluation for dialogue-level evaluations. We leave this as our future work. 

\section{Conclusion}
In this paper, we present \textsc{DEnsity}, a new learnable metric for open-domain dialogue systems. 
\textsc{DEnsity} evaluates a response by estimating its density on the distribution of human conversations.
Empirical results on multiple datasets demonstrate that our metric has a higher correlation with human evaluations than other metrics. 
We hope that \textsc{DEnsity}, a reliable and robust metric for evaluating dialogue systems, contributes to improving evaluation of natural language generation tasks.



\section*{Limitations}
Our proposed metric is mainly designed for a turn-level evaluation of dialogue systems. 
We recognize that our metric may not generalize to other evaluation scenarios directly, such as dialogue-level evaluation or human-chatbot interactive setups. 
As shown in Section~\ref{appendix:dialogue_level}, the easiest way to extend our metric to a multi-turn dialogue evaluation is by evaluating every turn in a dialogue individually, and then aggregating their scores.
However, as the dialogue-level evaluation is not considered during the development process of our metric, it is not clear whether such a simple extension would be applicable without a decrease in performance.
Nevertheless, as turn-level evaluation is a fundamental component to build a holistic evaluation framework for a dialogue, we believe that it is an important task to investigate better evaluation metrics for individual responses.

\section*{Ethics Statement}
All experiments are conducted on English datasets only, so the generalizability toward other languages is not verified. Besides, as the current automatic evaluation metrics, including ours, are imperfect, they may introduce unintended favor toward a certain type of responses. Future research should focus on detecting and mitigating such undesirable biases of learnable metrics.


\section*{Acknowledgements}
This work was supported by Institute for Information \& communications Technology Planning \& Evaluation(IITP) grant funded by the Korea government(MSIT) (No. 2020-0-00368, A Neural-Symbolic Model for Knowledge Acquisition and Inference Techniques) and the Challengeable Future Defense Technology Research and Development Program through the Agency For Defense Development(ADD) funded by the Defense Acquisition Program Administration(DAPA) in 2022(No. N04220080).

\bibliography{custom}

\begin{thebibliography}{49}
\expandafter\ifx\csname natexlab\endcsname\relax\def\natexlab#1{#1}\fi

\bibitem[{Bak and Oh(2020)}]{bak-oh-2020-speaker}
JinYeong Bak and Alice Oh. 2020.
\newblock \href {https://doi.org/10.18653/v1/2020.acl-main.568} {Speaker
  sensitive response evaluation model}.
\newblock In \emph{Proceedings of the 58th Annual Meeting of the Association
  for Computational Linguistics}, pages 6376--6385, Online. Association for
  Computational Linguistics.

\bibitem[{Banerjee and Lavie(2005)}]{banerjee-lavie-2005-meteor}
Satanjeev Banerjee and Alon Lavie. 2005.
\newblock \href {https://aclanthology.org/W05-0909} {{METEOR}: An automatic
  metric for {MT} evaluation with improved correlation with human judgments}.
\newblock In \emph{Proceedings of the {ACL} Workshop on Intrinsic and Extrinsic
  Evaluation Measures for Machine Translation and/or Summarization}, pages
  65--72, Ann Arbor, Michigan. Association for Computational Linguistics.

\bibitem[{Devlin et~al.(2019)Devlin, Chang, Lee, and
  Toutanova}]{devlin-etal-2019-bert}
Jacob Devlin, Ming-Wei Chang, Kenton Lee, and Kristina Toutanova. 2019.
\newblock \href {https://doi.org/10.18653/v1/N19-1423} {{BERT}: Pre-training of
  deep bidirectional transformers for language understanding}.
\newblock In \emph{Proceedings of the 2019 Conference of the North {A}merican
  Chapter of the Association for Computational Linguistics: Human Language
  Technologies, Volume 1 (Long and Short Papers)}, pages 4171--4186,
  Minneapolis, Minnesota. Association for Computational Linguistics.

\bibitem[{Dinan et~al.(2020)Dinan, Logacheva, Malykh, Miller, Shuster, Urbanek,
  Kiela, Szlam, Serban, Lowe et~al.}]{dinan2020second}
Emily Dinan, Varvara Logacheva, Valentin Malykh, Alexander Miller, Kurt
  Shuster, Jack Urbanek, Douwe Kiela, Arthur Szlam, Iulian Serban, Ryan Lowe,
  et~al. 2020.
\newblock The second conversational intelligence challenge (convai2).
\newblock In \emph{The NeurIPS'18 Competition}, pages 187--208. Springer.

\bibitem[{Gao et~al.(2021)Gao, Yao, and Chen}]{gao-etal-2021-simcse}
Tianyu Gao, Xingcheng Yao, and Danqi Chen. 2021.
\newblock \href {https://doi.org/10.18653/v1/2021.emnlp-main.552} {{S}im{CSE}:
  Simple contrastive learning of sentence embeddings}.
\newblock In \emph{Proceedings of the 2021 Conference on Empirical Methods in
  Natural Language Processing}, pages 6894--6910, Online and Punta Cana,
  Dominican Republic. Association for Computational Linguistics.

\bibitem[{Ghazarian et~al.(2019)Ghazarian, Wei, Galstyan, and
  Peng}]{ghazarian-etal-2019-better}
Sarik Ghazarian, Johnny Wei, Aram Galstyan, and Nanyun Peng. 2019.
\newblock \href {https://doi.org/10.18653/v1/W19-2310} {Better automatic
  evaluation of open-domain dialogue systems with contextualized embeddings}.
\newblock In \emph{Proceedings of the Workshop on Methods for Optimizing and
  Evaluating Neural Language Generation}, pages 82--89, Minneapolis, Minnesota.
  Association for Computational Linguistics.

\bibitem[{Gunel et~al.(2020)Gunel, Du, Conneau, and
  Stoyanov}]{gunel2020supervised}
Beliz Gunel, Jingfei Du, Alexis Conneau, and Veselin Stoyanov. 2020.
\newblock Supervised contrastive learning for pre-trained language model
  fine-tuning.
\newblock In \emph{International Conference on Learning Representations}.

\bibitem[{Gupta et~al.(2021)Gupta, Tsvetkov, and
  Bigham}]{gupta-etal-2021-synthesizing}
Prakhar Gupta, Yulia Tsvetkov, and Jeffrey Bigham. 2021.
\newblock \href {https://doi.org/10.18653/v1/2021.findings-acl.338}
  {Synthesizing adversarial negative responses for robust response ranking and
  evaluation}.
\newblock In \emph{Findings of the Association for Computational Linguistics:
  ACL-IJCNLP 2021}, pages 3867--3883, Online. Association for Computational
  Linguistics.

\bibitem[{Hendrycks and Gimpel(2016)}]{hendrycks2016baseline}
Dan Hendrycks and Kevin Gimpel. 2016.
\newblock A baseline for detecting misclassified and out-of-distribution
  examples in neural networks.
\newblock \emph{arXiv preprint arXiv:1610.02136}.

\bibitem[{Holtzman et~al.(2019)Holtzman, Buys, Du, Forbes, and
  Choi}]{holtzman2019curious}
Ari Holtzman, Jan Buys, Li~Du, Maxwell Forbes, and Yejin Choi. 2019.
\newblock The curious case of neural text degeneration.
\newblock In \emph{International Conference on Learning Representations}.

\bibitem[{Huang et~al.(2020)Huang, Ye, Qin, Lin, and Liang}]{huang2020grade}
Lishan Huang, Zheng Ye, Jinghui Qin, Liang Lin, and Xiaodan Liang. 2020.
\newblock \href {https://doi.org/10.18653/v1/2020.emnlp-main.742} {{GRADE}:
  Automatic graph-enhanced coherence metric for evaluating open-domain dialogue
  systems}.
\newblock In \emph{Proceedings of the 2020 Conference on Empirical Methods in
  Natural Language Processing (EMNLP)}, pages 9230--9240, Online. Association
  for Computational Linguistics.

\bibitem[{Khalid and Lee(2022)}]{khalid-lee-2022-explaining}
Baber Khalid and Sungjin Lee. 2022.
\newblock \href {https://doi.org/10.18653/v1/2022.naacl-main.430} {Explaining
  dialogue evaluation metrics using adversarial behavioral analysis}.
\newblock In \emph{Proceedings of the 2022 Conference of the North American
  Chapter of the Association for Computational Linguistics: Human Language
  Technologies}, pages 5871--5883, Seattle, United States. Association for
  Computational Linguistics.

\bibitem[{Khosla et~al.(2020)Khosla, Teterwak, Wang, Sarna, Tian, Isola,
  Maschinot, Liu, and Krishnan}]{scl}
Prannay Khosla, Piotr Teterwak, Chen Wang, Aaron Sarna, Yonglong Tian, Phillip
  Isola, Aaron Maschinot, Ce~Liu, and Dilip Krishnan. 2020.
\newblock Supervised contrastive learning.
\newblock \emph{Advances in Neural Information Processing Systems},
  33:18661--18673.

\bibitem[{Lee et~al.(2018)Lee, Lee, Lee, and Shin}]{lee2018simple}
Kimin Lee, Kibok Lee, Honglak Lee, and Jinwoo Shin. 2018.
\newblock A simple unified framework for detecting out-of-distribution samples
  and adversarial attacks.
\newblock \emph{Advances in neural information processing systems}, 31.

\bibitem[{Lee et~al.(2022)Lee, Park, Choi, and Choo}]{lee-etal-2022-pneg}
Nyoungwoo Lee, ChaeHun Park, Ho-Jin Choi, and Jaegul Choo. 2022.
\newblock \href {https://aclanthology.org/2022.emnlp-main.733} {Pneg:
  Prompt-based negative response generation for dialogue response selection
  task}.
\newblock In \emph{Proceedings of the 2022 Conference on Empirical Methods in
  Natural Language Processing}, pages 10692--10703, Abu Dhabi, United Arab
  Emirates. Association for Computational Linguistics.

\bibitem[{Li et~al.(2017)Li, Su, Shen, Li, Cao, and
  Niu}]{li-etal-2017-dailydialog}
Yanran Li, Hui Su, Xiaoyu Shen, Wenjie Li, Ziqiang Cao, and Shuzi Niu. 2017.
\newblock \href {https://aclanthology.org/I17-1099} {{D}aily{D}ialog: A
  manually labelled multi-turn dialogue dataset}.
\newblock In \emph{Proceedings of the Eighth International Joint Conference on
  Natural Language Processing (Volume 1: Long Papers)}, pages 986--995, Taipei,
  Taiwan. Asian Federation of Natural Language Processing.

\bibitem[{Li et~al.(2021)Li, Zhang, Fei, Feng, and Zhou}]{flowscore}
Zekang Li, Jinchao Zhang, Zhengcong Fei, Yang Feng, and Jie Zhou. 2021.
\newblock \href {https://doi.org/10.18653/v1/2021.acl-long.11} {Conversations
  are not flat: Modeling the dynamic information flow across dialogue
  utterances}.
\newblock In \emph{Proceedings of the 59th Annual Meeting of the Association
  for Computational Linguistics and the 11th International Joint Conference on
  Natural Language Processing (Volume 1: Long Papers)}, pages 128--138, Online.
  Association for Computational Linguistics.

\bibitem[{Liang et~al.(2018)Liang, Li, and Srikant}]{liang2018enhancing}
Shiyu Liang, Yixuan Li, and R~Srikant. 2018.
\newblock Enhancing the reliability of out-of-distribution image detection in
  neural networks.
\newblock In \emph{International Conference on Learning Representations}.

\bibitem[{Lin(2004)}]{lin-2004-rouge}
Chin-Yew Lin. 2004.
\newblock \href {https://aclanthology.org/W04-1013} {{ROUGE}: A package for
  automatic evaluation of summaries}.
\newblock In \emph{Text Summarization Branches Out}, pages 74--81, Barcelona,
  Spain. Association for Computational Linguistics.

\bibitem[{Liu et~al.(2016)Liu, Lowe, Serban, Noseworthy, Charlin, and
  Pineau}]{liu-etal-2016-evaluate}
Chia-Wei Liu, Ryan Lowe, Iulian Serban, Mike Noseworthy, Laurent Charlin, and
  Joelle Pineau. 2016.
\newblock \href {https://doi.org/10.18653/v1/D16-1230} {How {NOT} to evaluate
  your dialogue system: An empirical study of unsupervised evaluation metrics
  for dialogue response generation}.
\newblock In \emph{Proceedings of the 2016 Conference on Empirical Methods in
  Natural Language Processing}, pages 2122--2132, Austin, Texas. Association
  for Computational Linguistics.

\bibitem[{Loshchilov and Hutter(2018)}]{adamw}
Ilya Loshchilov and Frank Hutter. 2018.
\newblock Decoupled weight decay regularization.
\newblock In \emph{International Conference on Learning Representations}.

\bibitem[{Lowe et~al.(2017)Lowe, Noseworthy, Serban, Angelard-Gontier, Bengio,
  and Pineau}]{lowe-etal-2017-adem}
Ryan Lowe, Michael Noseworthy, Iulian~Vlad Serban, Nicolas Angelard-Gontier,
  Yoshua Bengio, and Joelle Pineau. 2017.
\newblock \href {https://doi.org/10.18653/v1/P17-1103} {Towards an automatic
  {T}uring test: Learning to evaluate dialogue responses}.
\newblock In \emph{Proceedings of the 55th Annual Meeting of the Association
  for Computational Linguistics (Volume 1: Long Papers)}, pages 1116--1126,
  Vancouver, Canada. Association for Computational Linguistics.

\bibitem[{Mehri and Eskenazi(2020{\natexlab{a}})}]{mehri2020fed}
Shikib Mehri and Maxine Eskenazi. 2020{\natexlab{a}}.
\newblock Unsupervised evaluation of interactive dialog with dialogpt.
\newblock In \emph{Proceedings of the 21th Annual Meeting of the Special
  Interest Group on Discourse and Dialogue}, pages 225--235.

\bibitem[{Mehri and Eskenazi(2020{\natexlab{b}})}]{mehri-eskenazi-2020-usr}
Shikib Mehri and Maxine Eskenazi. 2020{\natexlab{b}}.
\newblock \href {https://doi.org/10.18653/v1/2020.acl-main.64} {{USR}: An
  unsupervised and reference free evaluation metric for dialog generation}.
\newblock In \emph{Proceedings of the 58th Annual Meeting of the Association
  for Computational Linguistics}, pages 681--707, Online. Association for
  Computational Linguistics.

\bibitem[{Pang et~al.(2020)Pang, Nijkamp, Han, Zhou, Liu, and
  Tu}]{pang-etal-2020-towards}
Bo~Pang, Erik Nijkamp, Wenjuan Han, Linqi Zhou, Yixian Liu, and Kewei Tu. 2020.
\newblock \href {https://doi.org/10.18653/v1/2020.acl-main.333} {Towards
  holistic and automatic evaluation of open-domain dialogue generation}.
\newblock In \emph{Proceedings of the 58th Annual Meeting of the Association
  for Computational Linguistics}, pages 3619--3629, Online. Association for
  Computational Linguistics.

\bibitem[{Papineni et~al.(2002)Papineni, Roukos, Ward, and
  Zhu}]{papineni-etal-2002-bleu}
Kishore Papineni, Salim Roukos, Todd Ward, and Wei-Jing Zhu. 2002.
\newblock \href {https://doi.org/10.3115/1073083.1073135} {{B}leu: a method for
  automatic evaluation of machine translation}.
\newblock In \emph{Proceedings of the 40th Annual Meeting of the Association
  for Computational Linguistics}, pages 311--318, Philadelphia, Pennsylvania,
  USA. Association for Computational Linguistics.

\bibitem[{Park et~al.(2021)Park, Jang, Yang, and
  Park}]{park-etal-2021-generating}
ChaeHun Park, Eugene Jang, Wonsuk Yang, and Jong Park. 2021.
\newblock \href {https://doi.org/10.18653/v1/2021.naacl-main.120} {Generating
  negative samples by manipulating golden responses for unsupervised learning
  of a response evaluation model}.
\newblock In \emph{Proceedings of the 2021 Conference of the North American
  Chapter of the Association for Computational Linguistics: Human Language
  Technologies}, pages 1525--1534, Online. Association for Computational
  Linguistics.

\bibitem[{Paszke et~al.(2019)Paszke, Gross, Massa, Lerer, Bradbury, Chanan,
  Killeen, Lin, Gimelshein, Antiga et~al.}]{paszke2019pytorch}
Adam Paszke, Sam Gross, Francisco Massa, Adam Lerer, James Bradbury, Gregory
  Chanan, Trevor Killeen, Zeming Lin, Natalia Gimelshein, Luca Antiga, et~al.
  2019.
\newblock Pytorch: An imperative style, high-performance deep learning library.
\newblock \emph{Advances in neural information processing systems}, 32.

\bibitem[{Pennington et~al.(2014)Pennington, Socher, and
  Manning}]{pennington-etal-2014-glove}
Jeffrey Pennington, Richard Socher, and Christopher Manning. 2014.
\newblock \href {https://doi.org/10.3115/v1/D14-1162} {{G}lo{V}e: Global
  vectors for word representation}.
\newblock In \emph{Proceedings of the 2014 Conference on Empirical Methods in
  Natural Language Processing ({EMNLP})}, pages 1532--1543, Doha, Qatar.
  Association for Computational Linguistics.

\bibitem[{Phy et~al.(2020)Phy, Zhao, and Aizawa}]{phy2020deconstruct}
Vitou Phy, Yang Zhao, and Akiko Aizawa. 2020.
\newblock Deconstruct to reconstruct a configurable evaluation metric for
  open-domain dialogue systems.
\newblock In \emph{Proceedings of the 28th International Conference on
  Computational Linguistics}, pages 4164--4178.

\bibitem[{Pillutla et~al.(2021)Pillutla, Swayamdipta, Zellers, Thickstun,
  Welleck, Choi, and Harchaoui}]{pillutla2021mauve}
Krishna Pillutla, Swabha Swayamdipta, Rowan Zellers, John Thickstun, Sean
  Welleck, Yejin Choi, and Zaid Harchaoui. 2021.
\newblock Mauve: Measuring the gap between neural text and human text using
  divergence frontiers.
\newblock \emph{Advances in Neural Information Processing Systems},
  34:4816--4828.

\bibitem[{Radford et~al.(2019)Radford, Wu, Child, Luan, Amodei, and
  Sutskever}]{radford2019gpt2}
Alec Radford, Jeffrey Wu, Rewon Child, David Luan, Dario Amodei, and Ilya
  Sutskever. 2019.
\newblock Language models are unsupervised multitask learners.
\newblock \emph{OpenAI Blog}, 1(8):9.

\bibitem[{Sai et~al.(2020)Sai, Mohankumar, Arora, and
  Khapra}]{sai-etal-2020-deb}
Ananya~B. Sai, Akash~Kumar Mohankumar, Siddhartha Arora, and Mitesh~M. Khapra.
  2020.
\newblock \href {https://doi.org/10.1162/tacl_a_00347} {Improving dialog
  evaluation with a multi-reference adversarial dataset and large scale
  pretraining}.
\newblock \emph{Transactions of the Association for Computational Linguistics},
  8:810--827.

\bibitem[{See et~al.(2017)See, Liu, and Manning}]{see2017get}
Abigail See, Peter~J. Liu, and Christopher~D. Manning. 2017.
\newblock \href {https://doi.org/10.18653/v1/P17-1099} {Get to the point:
  Summarization with pointer-generator networks}.
\newblock In \emph{Proceedings of the 55th Annual Meeting of the Association
  for Computational Linguistics (Volume 1: Long Papers)}, pages 1073--1083,
  Vancouver, Canada. Association for Computational Linguistics.

\bibitem[{Sellam et~al.(2020)Sellam, Das, and Parikh}]{sellam-etal-2020-bleurt}
Thibault Sellam, Dipanjan Das, and Ankur Parikh. 2020.
\newblock \href {https://doi.org/10.18653/v1/2020.acl-main.704} {{BLEURT}:
  Learning robust metrics for text generation}.
\newblock In \emph{Proceedings of the 58th Annual Meeting of the Association
  for Computational Linguistics}, pages 7881--7892, Online. Association for
  Computational Linguistics.

\bibitem[{Sharma et~al.(2017)Sharma, El~Asri, Schulz, and
  Zumer}]{sharma2017nlgeval}
Shikhar Sharma, Layla El~Asri, Hannes Schulz, and Jeremie Zumer. 2017.
\newblock \href {http://arxiv.org/abs/1706.09799} {Relevance of unsupervised
  metrics in task-oriented dialogue for evaluating natural language
  generation}.
\newblock \emph{CoRR}, abs/1706.09799.

\bibitem[{Sinha et~al.(2020)Sinha, Parthasarathi, Wang, Lowe, Hamilton, and
  Pineau}]{maude}
Koustuv Sinha, Prasanna Parthasarathi, Jasmine Wang, Ryan Lowe, William~L.
  Hamilton, and Joelle Pineau. 2020.
\newblock \href {https://doi.org/10.18653/v1/2020.acl-main.220} {Learning an
  unreferenced metric for online dialogue evaluation}.
\newblock In \emph{Proceedings of the 58th Annual Meeting of the Association
  for Computational Linguistics}, pages 2430--2441, Online. Association for
  Computational Linguistics.

\bibitem[{Speer et~al.(2017)Speer, Chin, and Havasi}]{speer2017conceptnet}
Robyn Speer, Joshua Chin, and Catherine Havasi. 2017.
\newblock Conceptnet 5.5: An open multilingual graph of general knowledge.
\newblock In \emph{Thirty-first AAAI conference on artificial intelligence}.

\bibitem[{Sutskever et~al.(2014)Sutskever, Vinyals, and
  Le}]{sutskever2014sequence}
Ilya Sutskever, Oriol Vinyals, and Quoc~V Le. 2014.
\newblock Sequence to sequence learning with neural networks.
\newblock \emph{Advances in neural information processing systems}, 27.

\bibitem[{Tao et~al.(2018)Tao, Mou, Zhao, and Yan}]{tao2018ruber}
Chongyang Tao, Lili Mou, Dongyan Zhao, and Rui Yan. 2018.
\newblock Ruber: An unsupervised method for automatic evaluation of open-domain
  dialog systems.
\newblock In \emph{Thirty-Second AAAI Conference on Artificial Intelligence}.

\bibitem[{Vaswani et~al.(2017)Vaswani, Shazeer, Parmar, Uszkoreit, Jones,
  Gomez, Kaiser, and Polosukhin}]{vaswani2017attention}
Ashish Vaswani, Noam Shazeer, Niki Parmar, Jakob Uszkoreit, Llion Jones,
  Aidan~N Gomez, {\L}ukasz Kaiser, and Illia Polosukhin. 2017.
\newblock Attention is all you need.
\newblock \emph{Advances in neural information processing systems}, 30.

\bibitem[{Winkens et~al.(2020)Winkens, Bunel, Roy, Stanforth, Natarajan,
  Ledsam, MacWilliams, Kohli, Karthikesalingam, Kohl
  et~al.}]{winkens2020contrastive}
Jim Winkens, Rudy Bunel, Abhijit~Guha Roy, Robert Stanforth, Vivek Natarajan,
  Joseph~R Ledsam, Patricia MacWilliams, Pushmeet Kohli, Alan Karthikesalingam,
  Simon Kohl, et~al. 2020.
\newblock Contrastive training for improved out-of-distribution detection.
\newblock \emph{arXiv preprint arXiv:2007.05566}.

\bibitem[{Wolf et~al.(2020)Wolf, Debut, Sanh, Chaumond, Delangue, Moi, Cistac,
  Rault, Louf, Funtowicz, Davison, Shleifer, von Platen, Ma, Jernite, Plu, Xu,
  Scao, Gugger, Drame, Lhoest, and Rush}]{wolf-etal-2020-transformers}
Thomas Wolf, Lysandre Debut, Victor Sanh, Julien Chaumond, Clement Delangue,
  Anthony Moi, Pierric Cistac, Tim Rault, Rémi Louf, Morgan Funtowicz, Joe
  Davison, Sam Shleifer, Patrick von Platen, Clara Ma, Yacine Jernite, Julien
  Plu, Canwen Xu, Teven~Le Scao, Sylvain Gugger, Mariama Drame, Quentin Lhoest,
  and Alexander~M. Rush. 2020.
\newblock \href {https://www.aclweb.org/anthology/2020.emnlp-demos.6}
  {Transformers: State-of-the-art natural language processing}.
\newblock In \emph{Proceedings of the 2020 Conference on Empirical Methods in
  Natural Language Processing: System Demonstrations}, pages 38--45, Online.
  Association for Computational Linguistics.

\bibitem[{Xiang et~al.(2021)Xiang, Liu, Cai, Li, Lian, and
  Liu}]{xiang2021assessing}
Jiannan Xiang, Yahui Liu, Deng Cai, Huayang Li, Defu Lian, and Lemao Liu. 2021.
\newblock \href {https://doi.org/10.18653/v1/2021.findings-acl.193} {Assessing
  dialogue systems with distribution distances}.
\newblock In \emph{Findings of the Association for Computational Linguistics:
  ACL-IJCNLP 2021}, pages 2192--2198, Online. Association for Computational
  Linguistics.

\bibitem[{Xu et~al.(2020)Xu, He, Yan, Liu, Liu, and Xu}]{xu-etal-2020-deep}
Hong Xu, Keqing He, Yuanmeng Yan, Sihong Liu, Zijun Liu, and Weiran Xu. 2020.
\newblock \href {https://doi.org/10.18653/v1/2020.coling-main.125} {A deep
  generative distance-based classifier for out-of-domain detection with
  mahalanobis space}.
\newblock In \emph{Proceedings of the 28th International Conference on
  Computational Linguistics}, pages 1452--1460, Barcelona, Spain (Online).
  International Committee on Computational Linguistics.

\bibitem[{Zhang et~al.(2021)Zhang, Chen, D{'}Haro, Zhang, Friedrichs, Lee, and
  Li}]{zhang-etal-2021-dynaeval}
Chen Zhang, Yiming Chen, Luis~Fernando D{'}Haro, Yan Zhang, Thomas Friedrichs,
  Grandee Lee, and Haizhou Li. 2021.
\newblock \href {https://doi.org/10.18653/v1/2021.acl-long.441} {{D}yna{E}val:
  Unifying turn and dialogue level evaluation}.
\newblock In \emph{Proceedings of the 59th Annual Meeting of the Association
  for Computational Linguistics and the 11th International Joint Conference on
  Natural Language Processing (Volume 1: Long Papers)}, pages 5676--5689,
  Online. Association for Computational Linguistics.

\bibitem[{Zhang et~al.(2019)Zhang, Kishore, Wu, Weinberger, and
  Artzi}]{zhang2019bertscore}
Tianyi Zhang, Varsha Kishore, Felix Wu, Kilian~Q Weinberger, and Yoav Artzi.
  2019.
\newblock Bertscore: Evaluating text generation with bert.
\newblock In \emph{International Conference on Learning Representations}.

\bibitem[{Zhao et~al.(2020)Zhao, Lala, and Kawahara}]{zhao2020designing}
Tianyu Zhao, Divesh Lala, and Tatsuya Kawahara. 2020.
\newblock \href {https://doi.org/10.18653/v1/2020.acl-main.4} {Designing
  precise and robust dialogue response evaluators}.
\newblock In \emph{Proceedings of the 58th Annual Meeting of the Association
  for Computational Linguistics}, pages 26--33, Online. Association for
  Computational Linguistics.

\bibitem[{Zhou et~al.(2021)Zhou, Liu, and Chen}]{zhou2021contrastiveOOD}
Wenxuan Zhou, Fangyu Liu, and Muhao Chen. 2021.
\newblock \href {https://doi.org/10.18653/v1/2021.emnlp-main.84} {Contrastive
  out-of-distribution detection for pretrained transformers}.
\newblock In \emph{Proceedings of the 2021 Conference on Empirical Methods in
  Natural Language Processing}, pages 1100--1111, Online and Punta Cana,
  Dominican Republic. Association for Computational Linguistics.

\end{thebibliography}
\bibliographystyle{acl_natbib}

\appendix

\clearpage

\section*{Appendix}

\section{Baseline Details}
\label{appendix:baseline}
We present further implementation details in baseline metrics.
For \textbf{BLEU}, we use BLEU-2 score with NLTK library\footnote{\url{https://www.nltk.org/}}.
For \textbf{ROUGE}, we use F-score of ROUGE-L.
For \textbf{METEOR}, we use NLTK library\footnote{\url{https://www.nltk.org/}}.
For \textbf{Embedding Average/Greedy/Extrema}, we use an evaluation toolkit released by \citet{sharma2017nlgeval} with GloVe~\citep{pennington-etal-2014-glove} embedding.
For \textbf{BERTScore}, we use the default \url{roberta-large} model in the official implementation\footnote{\url{https://github.com/Tiiiger/bert_score}}, and use F1 score between answer and generated responses for evaluation. 
For \textbf{SimCSE}, we use \url{princeton-nlp/sup-simcse-bert-base-uncased} model in the Huggingface Hub\footnote{\url{https://huggingface.co/princeton-nlp/sup-simcse-bert-base-uncased}}, and compute the cosine similarity between answer and generated responses for evaluation. 
For \textbf{BLEURT}, we use \url{Elron/bleurt-tiny-512} model in the Huggingface Hub\footnote{\url{https://huggingface.co/Elron/bleurt-tiny-512}}.
For \textbf{USR-MLM}, we train \url{bert-} \url{base-uncased} model with learning rate, train epochs, and batch size as 5e-5 and 1, and 16, respectively.
For \textbf{GPT2-Coherency}, we train 12-layer \url{gpt2} model with learning rate, maximum train epochs, and batch size as 5e-5, 10, and 16, respectively.
For \textbf{BERT-RUBER}, we use \url{bert-base-uncased} as an contextualized word embedding, and train the model with learning rate, maximum train epoch, and batch size as 1e-4, 10, and 16, respectively.
For \textbf{USR-Retrieval}, we train \url{bert-base-uncased} model with learning rate, train epochs, and batch size as 5e-5, 10, and 16, respectively.
For \textbf{USL-H}, we use an official implementation\footnote{\url{https://github.com/vitouphy/usl_dialogue_metric}} to train and evaluate models.
For \textbf{FlowScore} and \textbf{DynaEval}, we use official models\footnote{\url{https://github.com/ictnlp/DialoFlow}}\footnote{\url{https://github.com/e0397123/DynaEval}} for evaluation.
All baseline models trained in our experiments utilize the same training environments with our metric~(e.g., optimizer, max sequence length) unless specified otherwise.
When evaluating reference-free models, the dialogue corpus for the training of models and the original dialogue corpus that derives an evaluation dataset are matched. 
For instance, we use DynaEval model trained on DailyDialog dataset for the evaluations on DailyDialog-Zhao and DailyDialog-GRADE datasets.

\section{Further Implementation Details}
\label{appendix:implementation}
We evaluate our selection model after every train epoch, and select the best model based on its recall@1 score on the validation set of the original dialogue corpus. 
All the learnable metrics are implemented with PyTorch~\citep{paszke2019pytorch}.
We use Transformers framework~\citep{wolf-etal-2020-transformers} from Huggingface\footnote{\url{https://huggingface.co/}} to implement transformer~\citep{vaswani2017attention}-based models.
In overall experiments, two 3090 RTX GPU with 24GB of memory are used.


\section{Additional Results}
\label{appendix:additional_results}

\begin{table}[t!]
\small
\begin{adjustbox}{width=\columnwidth}
\centering
\begin{tabular}{lcccc}
\toprule
 &  \multicolumn{2}{c}{\textbf{DailyDialog}}  &  \multicolumn{2}{c}{\textbf{ConvAI2}}   \\
$\textbf{Model}$ & R@1 & MRR & R@1 & MRR  \\
\cline{0-4}\noalign{\vskip 0.75ex}
 $\mathcal{L}_{RS}$ & 88.8 & 93.38 & 85.16 & 90.97   \\
 $\mathcal{L}_{RS}+\mathcal{L}_{CL}$ & 90.62 & 93.96 & 85.99 & 91.48 \\

 \bottomrule
\end{tabular}
\end{adjustbox}
\caption{Performance of response selection tasks on DailyDialog and ConvAI2 datasets. $\mathcal{L}_{RS}$ denotes the model trained with response selection task, while $\mathcal{L}_{RS}+\mathcal{L}_{CL}$ further utilizes contrastive learning for training.}
\label{tab:selection_performance}
\end{table}

\subsection{Impacts of Contrastive Learning on Response Selection Task}
To more comprehensively understand the effect of contrastive learning on our feature extractor~($g$), we report the impacts of our contrastive learning objective on the performance of the response selection task.
Specifically, we compare the selection model trained with both the response selection task and contrastive learning~($\mathcal{L}_{RS}+\mathcal{L}_{CL}$) against the model trained solely with response selection task~($\mathcal{L}_{RS}$). Recall@1 and mean reciprocal rank~(MRR) metrics are used to measure the selection accuracy. 
We evaluate both models on the test splits of DailyDialog and ConvAI2 datasets.
As shown in Table~\ref{tab:selection_performance}, we observe that contrastive learning improves the performance on the original task on which the selection models are trained.


\end{document}